\def\paperID{5711}
\def\confName{CVPR}
\def\confYear{2024}

\def\paperTitle{pixelSplat: 3D Gaussian Splats from Image Pairs\\for Scalable Generalizable 3D Reconstruction}

\def\authorBlock{
    David Charatan\textsuperscript{1} \qquad
    Sizhe Lester Li\textsuperscript{1} \qquad
    Andrea Tagliasacchi\textsuperscript{2} \qquad
    Vincent Sitzmann\textsuperscript{1} \\
    \textsuperscript{1}Massachusetts Institute of Technology \qquad
    \textsuperscript{2}Simon Fraser University, University of Toronto\\
    {\tt\small \{charatan, sizheli, sitzmann\}@mit.edu \qquad andrea.tagliasacchi@sfu.ca}
}

\newif\ifreview 
\newif\ifarxiv \newcommand{\arxiv}{\arxivtrue}
\newif\ifcamera \newcommand{\cameraready}{\cameratrue}
\newif\ifrebuttal

\arxiv
\cameraready

\pdfoutput=1
\documentclass[10pt,twocolumn,letterpaper]{article}
\ifreview \usepackage[review]{cvpr} \fi
\ifarxiv \usepackage[pagenumbers]{cvpr} \fi
\ifrebuttal \usepackage[rebuttal]{cvpr} \fi
\ifcamera \usepackage{cvpr} \fi

\usepackage{graphicx}	
\usepackage{amsmath}	
\usepackage{amssymb}	
\usepackage{booktabs}
\usepackage{times}
\usepackage{microtype}
\usepackage{epsfig}
\usepackage[table,xcdraw,dvipsnames]{xcolor}
\usepackage{caption}
\usepackage{float}
\usepackage{placeins}
\usepackage{color, colortbl}
\usepackage{stfloats}
\usepackage{enumitem}
\usepackage{tabularx}
\usepackage{xstring}
\usepackage{multirow}
\usepackage{xspace}
\usepackage{url}
\usepackage{subcaption}
\usepackage{xcolor}
\usepackage[hang,flushmargin]{footmisc}
\usepackage{layouts}
\usepackage{svg}
\usepackage{algorithm}
\usepackage{algpseudocode}
\usepackage{siunitx}
\usepackage{amsfonts}    %
\usepackage{paralist}
\usepackage{comment}
\usepackage{wrapfig}

\ifcamera \usepackage[accsupp]{axessibility} \fi

\ifarxiv  \fi

\definecolor{MyGreen}{rgb}{0, 0.55, 0}

\renewcommand{\paragraph}[1]{\vspace{.5em}\noindent\textbf{#1}.}

\def\loss{\mathcal{L}}

\def\depthprobs{\P}

\def\sphhar{\mathbf{S}}
\def\depthbins{\mathbf{b}}
\def\featmap{\mathbf{F}}
\def\origin{\mathbf{o}}
\def\direction{\mathbf{d}}
\def\pose{\mathbf{T}}
\def\ints{\mathbf{K}}
\def\depth{d}
\def\cov{\boldsymbol{\Sigma}}
\def\density{\boldsymbol{\alpha}}

\def\offset{\boldsymbol{\delta}}

\renewcommand{\P}{{\boldsymbol{\phi}}}
\def\loss{\mathcal{L}}
\def\mean{\boldsymbol{\mu}}

\def\pixcoord{\mathbf{u}}

\def\Q{\mathbf{Q}}
\def\K{\mathbf{K}}
\def\V{\mathbf{V}}

\newcommand{\img}{\textbf{I}}

\newcommand{\gausspars}{\mathbf{g}}
\newcommand{\plusequal}{\mathrel{+}=}

\newcommand{\R}[1]{{%
    \textbf{%
        \ifstrequal{#1}{1}{\textcolor{red}{R#1}}{%
        \ifstrequal{#1}{2}{\textcolor{blue}{R#1}}{%
        \ifstrequal{#1}{3}{\textcolor{magenta}{R#1}}{%
        \ifstrequal{#1}{4}{\textcolor{teal}{R#1}}{%
                           \textcolor{cyan}{R#1}%
        }}}}%
    }%
}}

\definecolor{tabfirst}{rgb}{1, 0.7, 0.7} %
\definecolor{tabsecond}{rgb}{1, 0.85, 0.7} %
\definecolor{tabthird}{rgb}{1, 1, 0.7} %

\definecolor{darkyellow}{rgb}{0.19,0.8,0.59} %

\usepackage{xr-hyper}

\makeatletter
\newcommand*{\addFileDependency}[1]{
  \typeout{(#1)}
  \@addtofilelist{#1}
  \IfFileExists{#1}{}{\typeout{No file #1.}}
}

\makeatother

\definecolor{cvprblue}{rgb}{0.21,0.49,0.74}
\usepackage[pagebackref,breaklinks,colorlinks,citecolor=cvprblue]{hyperref}
\usepackage[capitalize]{cleveref}
\crefname{section}{Sec.}{Secs.}
\crefname{table}{Table}{Tables}
\crefname{figure}{Fig.}{Figs.}

\begin{document}
\title{\paperTitle}

\author{\authorBlock}
\maketitle

\begin{abstract}

We introduce pixelSplat, a feed-forward model that learns to reconstruct 3D radiance fields parameterized by 3D Gaussian primitives from pairs of images.
Our model features real-time and memory-efficient rendering for scalable training as well as fast 3D reconstruction at inference time.
To overcome local minima inherent to sparse and locally supported representations, we predict a dense probability distribution over~3D and sample Gaussian means from that probability distribution.
We make this sampling operation differentiable via a reparameterization trick, allowing us to back-propagate gradients through the Gaussian splatting representation.
We benchmark our method on wide-baseline novel view synthesis on the real-world RealEstate10k and ACID datasets, where we outperform state-of-the-art light field transformers and accelerate rendering by 2.5 orders of magnitude while reconstructing an interpretable and editable 3D radiance field. Additional materials can be found on the project website. \footnote{\url{dcharatan.github.io/pixelsplat}}
\end{abstract}

\section{Introduction}
\label{sec:intro}
We investigate the problem of generalizable novel view synthesis from sparse image observations. 
This line of work has been revolutionized by differentiable rendering~\cite{mildenhall2020nerf,sitzmann2019deepvoxels,sitzmann2019scene,tewari2022advances} but has also inherited its key weakness: training, reconstruction, and rendering are notoriously memory- and time-intensive because differentiable rendering requires evaluating dozens or hundreds of points along each camera ray~\cite{pixelnerf}.

This has motivated light-field transformers~\cite{suhail2022light, du2023cross, sajjadi2021scene, sitzmann2021light}, where a ray is rendered by embedding it into a query token and a color is obtained via cross-attention over image tokens. While significantly faster than volume rendering, such methods are still far from real-time.
Additionally, they do not reconstruct 3D scene representations that can be edited or exported for downstream tasks in vision and graphics.

Meanwhile, recent work on single-scene novel view synthesis has shown that it is possible to use 3D Gaussian primitives to enable real-time rendering with little memory cost via rasterization-based volume rendering~\cite{kerbl20233d}.

We present pixelSplat, which brings the benefits of a primitive-based 3D representation---fast and memory-efficient rendering as well as interpretable 3D structure---to generalizable view synthesis. 
This is no straightforward task. First, in real-world datasets, camera poses are only reconstructed up to an arbitrary scale factor. We address this by designing a multi-view epipolar transformer that reliably infers this per-scene scale factor.
Next, optimizing primitive parameters directly via gradient descent suffers from local minima. In the single-scene case, this can be addressed via non-differentiable pruning and division heuristics~\cite{kerbl20233d}.
In contrast, in the generalizable case, we need to back-propagate gradients through the representation and thus cannot rely on non-differentiable operations.
\begin{figure}[t]
    \centering
    \includegraphics[width=\linewidth]{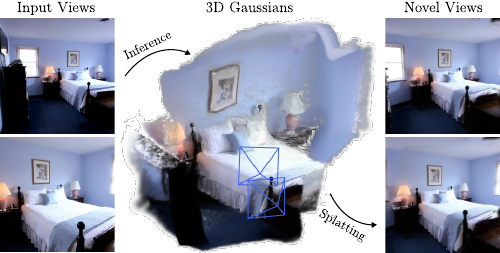}
    \vspace{-10pt}
    \caption{\textbf{Overview.} Given a pair of input images, pixelSplat reconstructs a 3D radiance field parameterized via 3D Gaussian primitives. This yields an explicit 3D representation that is renderable in real time, remains editable, and is cheap to train.}
    \vspace{-12pt}
    \label{fig:teaser}
\end{figure}

We thus propose a method by which Gaussian primitives can implicitly be spawned or deleted during training, avoiding local minima, but which nevertheless maintains gradient flow.
Specifically, we parameterize the positions (i.e., means) of Gaussians \emph{implicitly} via dense probability distributions predicted by our encoder. In each forward pass, we sample Gaussian primitive locations from this distribution.
We make the sampling operation differentiable via a reparameterization trick that couples the density of a sampled Gaussian primitive to the probability of that location.
When receiving a gradient that would increase the opacity of a Gaussian at a 3D location, our model increases the probability that the Gaussian will be sampled at that location again in the future.

We demonstrate the efficacy of our method by showcasing, for the first time, how a 3D Gaussian splatting representation can be predicted in a \textit{single forward pass} from just a pair of images.
In other words, we demonstrate how 3D Gaussians can be integrated in an end-to-end differentiable system.
We significantly outperform previous black-box based light field transformers on the real-world ACID and RealEstate10k datasets while drastically reducing both training and rendering cost and generating explicit 3D scenes. 

\section{Related Work}
\label{sec:related}

\paragraph{Single-scene novel view synthesis}
Advancements in neural rendering~\cite{tewari2022advances} and neural fields~\cite{xie2021neural,sitzmann2019siren,mildenhall2020nerf} have revolutionized 3D reconstruction and novel view synthesis from collections of posed images.
Recent approaches generally create 3D scene representations by backpropagating image-space photometric error through differentiable renderers. 
Early methods employed voxel grids and learned rendering techniques~\cite{nguyen2018rendernet,sitzmann2019deepvoxels,lombardi2019neural}. More recently, neural fields~\cite{xie2021neural,mildenhall2020nerf,barron2021mipnerf,martin2021nerf} and volume rendering~\cite{tagliasacchi2022volume,mildenhall2020nerf,lombardi2019neural} have become the de-facto standard. 
However, a key hurdle of these methods is their high computational demand, as rendering usually requires dozens of queries of the neural field per ray. 
Discrete data structures can accelerate rendering~\cite{mueller2022instant,fridovich2022plenoxels,chen2022tensorf,liu2020neural} but fall short of real-time rendering at high resolutions. 
3D Gaussian splatting~\cite{kerbl20233d} solves this problem by representing the radiance field using 3D Gaussians that can efficiently be rendered via rasterization.
However, all single-scene optimization methods require dozens of images to achieve high-quality novel view synthesis.
In this work, we train neural networks to estimate the parameters of a 3D Gaussian primitive scene representation from just two images in a single forward pass.

\paragraph{Prior-based 3D Reconstruction and View Synthesis}
Generalizable novel view synthesis seeks to enable 3D reconstruction and novel view synthesis from only a handful of images per scene.
If proxy geometry (e.g., depth maps) is available, machine learning can be combined with image-based rendering~\cite{Riegler2020FVS,aliev2020neural,kopanas2021point,wiles2020synsin} to produce convincing results.
Neural networks can also be trained to directly regress multi-plane images for small-baseline novel view synthesis~\cite{realestate10k,srinivasan2019pushing,tucker2020single,zhang2022video}. 
Large-baseline novel view synthesis, however, requires full 3D representations. 
Early approaches based on neural fields~\cite{sitzmann2019scene, niemeyer2020differentiable} encoded 3D scenes in individual latent codes and were thus limited to single-object scenes. 
Preserving end-to-end locality and shift equivariance between encoder and scene representation via pixel-aligned features~\cite{pixelnerf,lin2022vision,trevithick2021grf,sharma2022neural,guo2022fast} or via transformers~\cite{wang2021ibrnet,reizenstein2021common} has enabled generalization to unbounded scenes.
Inspired by classical multi-view stereo, neural networks have also been combined with cost volumes to match features across views~\cite{chen2021mvsnerf,liu2022neural,johari2022geonerf,SRF}.
While the above methods infer interpretable 3D representations in the form of signed distances or radiance fields, recent light field scene representations trade interpretability for faster rendering~\cite{sitzmann2021light,suhail2022light,du2023cross,suhail2022generalizable,sajjadi2021scene}. 
Our method presents the best of both worlds: it infers an interpretable 3D scene representation in the form of 3D Gaussians while accelerating rendering by three orders of magnitude compared to light field transformers.

\paragraph{Scale ambiguity in machine learning for multi-view geometry}
Prior work has recognized the challenge of scene scale ambiguity.
In monocular depth estimation, state-of-the-art models rely on sophisticated scale-invariant depth losses~\cite{ranftl2020towards,ranftl2021vision,eigen2014depth,godard2019digging}.
In novel view synthesis, recent single-image 3D diffusion models trained on real-world data rescale 3D scenes according to heuristics on depth statistics and condition their encoders on scene scale~\cite{tewari2023diffusion,sargent2023zeronvs,chan2023generative}.
In this work, we instead build a multi-view encoder that can infer the scale of the scene.
We accomplish this using an epipolar transformer that finds cross-view pixel correspondences and associates them with positionally encoded depth values~\cite{epipolartransformers}.

\section{Background: 3D Gaussian Splatting}
\label{sec:background}
3D Gaussian Splatting~\cite{kerbl20233d}, which we will refer to as ~3D\nobreakdash-GS, parameterizes a 3D scene as a set of 3D Gaussian primitives~$\{\gausspars_k {=} (\mean_k, \cov_k, \density_k, \sphhar_k) \}_{k}^K$ which each have a mean~$\mean_k$, a covariance~$\cov_k$, an opacity ~$\density_k$, and spherical harmonics coefficients~$\sphhar_k$.
These primitives parameterize the 3D radiance field of the underlying scene and can be rendered to produce novel views.
However, unlike dense representations like neural fields~\cite{mildenhall2020nerf} and voxel grids~\cite{fridovich2022plenoxels}, Gaussian primitives can be rendered via an inexpensive rasterization operation~\cite{ranftl2020towards}.
Compared to the sampling-based approach used to render dense fields, this approach is significantly cheaper in terms of time and memory.

\paragraph{Local minima} 
A key challenge of function fitting with primitives is their susceptibility to local minima.
The fitting of a 3D-GS model is closely related to the fitting of a Gaussian mixture model, where we seek the parameters of a set of Gaussians such that we maximize the likelihood of a set of samples.
This problem is famously non-convex and generally solved with the Expectation-Maximization (EM) algorithm~\cite{dempster1977maximum}.
However, the EM algorithm still suffers from local minima~\cite{jin2016local} and is not applicable to inverse graphics, where only images of the 3D scene are provided and not ground-truth 3D volume density.
In 3D-GS, local minima arise when Gaussian primitives initialized at random locations have to move through space to arrive at their final location.
Two issues prevent this: first, Gaussian primitives have local support, meaning that gradients vanish if the distance to the correct location exceeds more than a few standard deviations.
Second, even if a Gaussian is close enough to a ``correct'' location to receive substantial gradients, there still needs to exist a path to its final location along which loss decreases monotonically.
In the context of differentiable rendering, this is generally not the case, as Gaussians often have to traverse empty space where they may occlude background features.
3D-GS relies on non-differentiable pruning and splitting operations dubbed ``Adaptive Density Control'' (see Sec.~5 of \cite{kerbl20233d}) to address this problem. 
However, these techniques are incompatible with the generalizable setting, where primitive parameters are predicted by a neural network that must receive gradients.
In section~\ref{sec:probabilistic_depth_prediction}, we thus propose a differentiable parameterization of Gaussian primitives that is not susceptible to local minima, allowing its use as a potential building block in larger end-to-end differentiable models. 

\section{Image-conditioned 3D Gaussian Inference}
\label{sec:method}

We present pixelSplat, a Gaussian-based generalizable novel view synthesis model.
Given a pair of images and their associated camera parameters, our method infers a 3D Gaussian representation of the underlying scene, which can be rendered to produce images of unseen viewpoints.
Our method consists of a two-view image encoder and a pixel-aligned Gaussian prediction module.
\Cref{sec:method-image-encoding} introduces our image encoding scheme.
We show that our image encoding scheme addresses scale ambiguity, which is a key challenge during Gaussian inference from real-world images.
\Cref{sec:probabilistic_depth_prediction} describes how our model predicts 3D Gaussian parameters, emphasizing how it overcomes the issue of local minima described in~\Cref{sec:background}.

\subsection{Resolving Scale Ambiguity}
\label{sec:method-image-encoding}
\begin{figure}[hbt]
    \vspace{-1em}
    \centering
    \includegraphics[width=0.9\linewidth]{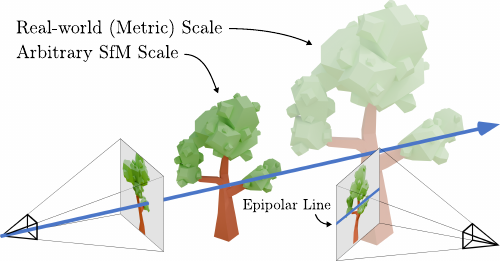}
    \vspace{-5pt}
    \caption{\textbf{Scale ambiguity.} SfM does not reconstruct camera poses in real-world, metric scale---poses are scaled by an arbitrary scale factor that is different for each scene. To render correct views, our model's 3D reconstruction needs to be consistent with this arbitrary scale. We illustrate how our epipolar encoder solves this problem. Features belonging to the ray's corresponding pixel on the left are compared with features sampled along the epipolar line on the right. Epipolar samples are augmented with their positionally-encoded depths along the ray, which allows our encoder to record correct depths. Recorded depths are later used for depth prediction.}
    \vspace{-15pt}
    \label{fig:scale_ambiguity}
\end{figure}

In an ideal world, novel-view-synthesis datasets would contain camera poses that are \emph{metric}.
If this were the case, each scene $\mathcal{C}^\text{m}_i$ would consist of a series of tuples $\mathcal{C}^\text{m}_i = \{ (\img_j, \pose^\text{m}_j) \}_j$ containing images $\img_j$ and corresponding real-world-scale poses $\pose^\text{m}_j$.
However, in practice, such datasets provide poses that are computed using structure-from-motion (SfM) software.
SfM reconstructs each scene \emph{only up to scale}, meaning that different scenes $\mathcal{C}_i$ are scaled by individual, arbitrary scale factors $s_i$.
A given scene $\mathcal{C}_i$ thus provides $\mathcal{C}_i = \{ (\img_j, s_i\pose^\text{m}_j) \}_{j}$, where $s_i\pose^\text{m}_j$ denotes a metric pose whose translation component is scaled by the unknown scalar $s_i \in \mathbb{R}^+$.
Note that recovering $s_i$ from a single image is \emph{impossible} due to the principle of scale ambiguity.
In practice, this means that a neural network making predictions about the geometry of a scene from a single image \emph{cannot possibly} predict the depth that matches the poses reconstructed by structure-from-motion. 
In monocular depth estimation, this has been addressed via scale-invariant losses~\cite{eigen2014depth,godard2019digging,ranftl2020towards}.
Our encoder similarly has to predict the geometry of the scene, chiefly via the position of each Gaussian primitive, which depends on the per-scene scale~$s_i$.
Refer to~\Cref{fig:scale_ambiguity} for an illustration of this challenge.

We thus propose a two-view encoder to resolve scale ambiguity and demonstrate in our ablations~(\Cref{tab:ablations}) that this is absolutely critical for 3D-structured novel view synthesis.
Let us denote the two reference views as $\img$ and $\tilde\img$.
Intuitively, for each pixel in $\img$, we will annotate points along its epipolar line in $\tilde\img$ with their corresponding depths in $\img$'s coordinate frame. Note that these depth values are computed from $\img$ and~$\tilde\img$'s camera poses, and thus encode the scene's scale $s_i$. Our encoder then finds per-pixel correspondence via epipolar attention~\cite{epipolartransformers} and memorizes the corresponding depth for that pixel.
The depths of pixels without correspondences in $\tilde\img$ are in-painted via per-image self-attention.
In the following paragraphs, we discuss these steps in detail.

We first encode each view separately into feature volumes $\featmap$ and $\tilde\featmap$ via a per-image feature encoder. 
Let $\pixcoord$ be pixel coordinates from $\img$, and $\ell$ be the epipolar line induced by its ray in~$\tilde\img$, i.e., the projection of $\pixcoord$'s camera ray onto the image plane of $\tilde\img$.
Along $\ell$, we now sample pixel coordinates~$\{\tilde{\pixcoord}_l\} \sim \tilde\img$. For each epipolar line sample $\tilde\pixcoord_l$, we further compute its distance to $\img$'s camera origin $\tilde\depth_{\tilde\pixcoord_l}$ by triangulation of $\pixcoord$ and $\tilde\pixcoord_l$. 
We then compute queries, keys and values for epipolar attention as follows:
\setlength{\abovedisplayskip}{8pt}
\setlength{\belowdisplayskip}{8pt}
\begin{gather}
\mathbf{s} =  \tilde\featmap[\tilde{\pixcoord}_l] \oplus \gamma(\tilde\depth_{\tilde\pixcoord_l}) \\
\mathbf{q} = \Q \cdot \featmap[\pixcoord], \quad \mathbf{k}_l = \K \cdot  \mathbf{s}, \quad \mathbf{v}_l = \V \cdot \mathbf{s},
\label{eq:pos_encoding}
\end{gather}
where, $\oplus$ denotes concatenation, $\gamma(\cdot)$ denotes positional encoding, and $\Q$, $\K$ and $\V$ are query, key and value embedding matrices.
We now perform epipolar cross-attention and update per-pixel features $\featmap[\pixcoord]$ as:
\begin{equation}
    \featmap[\pixcoord] \plusequal \text{Att}
    (\mathbf{q}, \{\mathbf{k}_l\}, \{\mathbf{v}_l\}),
\end{equation}
where $\text{Att}$ denotes softmax attention, and $\plusequal$ denotes a skip connection.
After this epipolar attention layer, each pixel feature $\featmap[\pixcoord]$ contains a weighted sum of the depth positional encodings, where we expect (and experimentally confirm in Sec. \ref{sect:analysis}) that the correct correspondence gained the largest weight, and that thus, each pixel feature $\featmap[\pixcoord]$ now encodes the scaled depth that is consistent with the arbitrary scale factor $s_i$ of the camera poses.
This epipolar cross-attention layer is followed by a per-image self-attention layer,
\begin{equation}
    \featmap \plusequal \text{SelfAttention}(\featmap).
\end{equation}
This enables our encoder to propagate scaled depth estimates to parts of the image feature maps that may not have any epipolar correspondences in the opposite image.

Note that this mechanism can be extended to more than two input views. See the supplemental material for details.

\begin{figure*}[tp]
    \centering
    \includegraphics[width=1.019\linewidth]{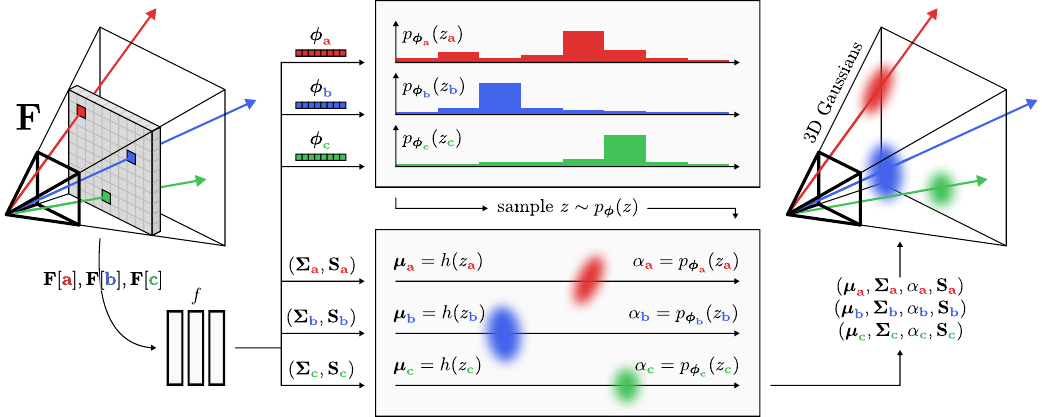}
    \caption{\textbf{Proposed probabilistic prediction of pixel-aligned Gaussians.}
    For every pixel feature $\featmap[\pixcoord]$ in the input feature map, a neural network $f$ predicts Gaussian primitive parameters $\cov$ and $\sphhar$. 
    Gaussian locations $\boldsymbol\mu$ and opacities $\density$ are not predicted directly, which would lead to local minima.
    Instead, $f$ predicts per-pixel discrete probability distributions over depths $p_{\depthprobs}(z)$, parameterized by $\depthprobs$.
    Sampling then yields the locations of Gaussian primitives. The opacity of each Gaussian is set to the probability of the sampled depth bucket. 
    The final set of Gaussian primitives can then be rendered from novel views using the splatting algorithm proposed by Kerbl et al.~\cite{kerbl20233d}.
   Note that for brevity, we use $h$ to represent the function that computes depths from bucket indices (see equations~\ref{eqn:bz}~and~\ref{eq:probdepth}).
    }
    \label{fig:overview}
    \vspace{-1em}
\end{figure*}

\begin{figure*}[t]
    \centering
    \includegraphics[width=\linewidth]{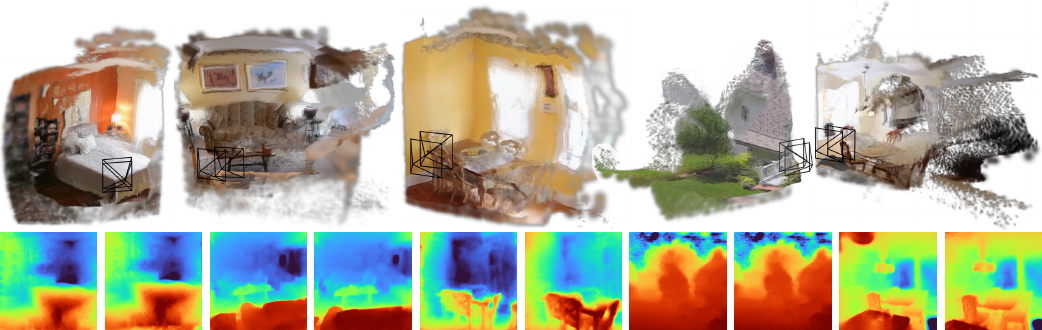}
    \caption{\textbf{3D Gaussians (top) and corresponding depth maps (bottom) predicted by our method.} In contrast to light field rendering methods like GPNR~\cite{suhail2022light} and that of Du et al.~\cite{du2023cross}, our method produces an \emph{explicit} 3D representation. Here, we show zoomed-out views of the Gaussians our method produces along with rendered depth maps, as viewed from the two reference viewpoints.}
    \label{fig:point_clouds}
\end{figure*} 

\begin{algorithm*}[b]
\caption{Probabilistic Prediction of a Pixel-Aligned Gaussian.}
\label{alg}
\begin{algorithmic}[1]
\Require Depth buckets $\depthbins \in \mathbb{R}^Z$, feature $\featmap[\pixcoord]$ at pixel coordinate $\pixcoord$, camera origin of reference view $\origin$, ray direction $\direction_{\pixcoord}$.
\State  $(\depthprobs, \offset, \cov, \sphhar) = f(\featmap[\pixcoord])$ 
\Comment{predict depth probabilities $\depthprobs$ and offsets $\offset$, covariance $\cov$, spherical harmonics coefficients $\sphhar$}
\State  $z \sim p_{\P}(z)$ \Comment{Sample depth bucket index $z$ from discrete probability distribution parameterized by $\depthprobs$}
\State  $\mean = \origin + (\depthbins_z + \offset_z) \direction_{\pixcoord}$ \Comment{Compute Gaussian mean $\mean$ by unprojecting with depth $\depthbins_z$ adjusted by bucket offset $\offset_z$}
\State  $\density = \depthprobs_z$ \Comment{Set Gaussian opacity $\density$ according to probability of sampled depth (\cref{par:reparameterization}).}
\State \Return $(\mean, \cov, \density, \sphhar)$
\end{algorithmic}
\end{algorithm*}

\subsection{Gaussian Parameter Prediction}
\label{sec:probabilistic_depth_prediction}
In this step, we leverage our scale-aware feature maps to predict the parameters of a set of Gaussian primitives~$\{\gausspars_k = (\mean_k, \cov_k, \density_k, \sphhar_k) \}_{k}^K$ that parameterize the scene's radiance field.
Every pixel in an image samples a point on a surface in the 3D scene.
We thus choose to parameterize the scene via \emph{pixel-aligned} Gaussians: for each pixel at coordinate $\pixcoord$, we take the corresponding feature $\featmap[\pixcoord]$ as input and predict the parameters of $M$ Gaussian primitives.
Note that we predict Gaussians from both reference views---the final set of Gaussian primitives is simply the union of the Gaussians predicted for each image.
In the following section, we discuss the case of $M{=}1$ primitives per pixel for simplicity; hence we aim to predict one tuple of values~$(\mean, \cov, \density, \sphhar)$ per pixel.
As we will see, the most consequential question is how to parameterize the position $\mean$ of each Gaussian.
Figure~\ref{fig:overview} illustrates the process by which we predict parameters of pixel-aligned Gaussians from a single reference image.

\paragraph{Baseline: predicting a point estimate of $\mean$} 
We first consider directly regressing a Gaussian's center~$\mean$.
This means implementing a neural network $g$ that regresses the distance $\depth \in \mathbb{R}^+$ of the Gaussian's mean from the camera origin, and then unprojecting it to 3D:
\begin{equation}
\mean = \origin + \depth_{\pixcoord} \: \direction,
\:\:
\depth = g(\featmap[\pixcoord]),
\:\:
\direction = \pose \ints^{-1} [\pixcoord, 1]^T
\label{eq:regress}
\end{equation}
where the ray direction $\direction$ is computed from the camera extrinsics $\mathbf{T}$ and intrinsics $\mathbf{K}$.
Unfortunately, directly optimizing Gaussian parameters is susceptible to local minima, as discussed in~\Cref{sec:background}.
Further, as we are back-propagating through the representation, we cannot rely on the spawning and pruning heuristics proposed in 3D-GS, as they are not differentiable.
Instead, we propose a differentiable alternative that succeeds in overcoming local minima.
In Table~\ref{tab:ablations}, we demonstrate that this leads to ${\approx} 1$dB of PSNR performance boost.

\paragraph{Proposed: predicting a probability density of $\mean$} 
Rather than predicting the depth~$\depth$ of a Gaussian, we predict the probability distribution of the likelihood that a Gaussian exists at a depth $\depth$ along the ray~$\pixcoord$.
We implement this as a discrete probability density over a set of depth buckets.
We set near and far planes $\depth_\text{near}$ and~$\depth_\text{far}$. Between these, we discretize depth into $Z$ bins, represented by a vector $\depthbins \in \mathbb{R}^Z$, where its $z$-th element $\depthbins_z$ is defined in disparity space as
\begin{equation}
    \depthbins_z = \left(\left(1 - \frac{z}{Z} \right) \left(\frac{1}{d_\text{near}} - \frac{1}{d_\text{far}}\right) + \frac{1}{d_\text{far}} \right)^{-1}.
    \label{eqn:bz}
\end{equation}
We then define a discrete probability distribution $p_\P(z)$ over an index variable $z$, parameterized by a vector of discrete probabilities $\P$, where its $z$-th element $\P_z$ is the probability that a surface exists in depth bucket $\depthbins_z$. 
Probabilities $\P$ are predicted by a fully connected neural network $f$ from per-pixel feature $\featmap[\pixcoord]$ at pixel coordinate $\pixcoord$ and normalized to sum to one via a softmax. 
Further, we predict a per-bucket center offset $\offset\in [0,1]^Z$ that adjusts the depth of the pixel-aligned Gaussian between bucket boundaries:
\begin{equation}
\mean = \origin + (\depthbins_z + \offset_z) \: \direction_{\pixcoord},
\:\:
z \sim p_\P(z), 
\:\:
(\P, \offset) = f(\featmap[\pixcoord])
\label{eq:probdepth}
\end{equation}
We note that, different from \cref{eq:regress}, during a forward pass a Gaussian location is \textit{sampled} from the distribution $z {\sim} p_\P(z)$, and the network predicts the \emph{probabilities} $\P$ instead of the depth $\depth$ directly.

\paragraph{Making sampling differentiable by setting $\density = \P_z$}
\label{par:reparameterization}
To train our model, we need to backpropagate gradients into the probabilities of the depth buckets $\P$. This means we need to compute the derivatives $\nabla_\P \mean$ of the mean $\mean$ with respect to probabilities $\P$.
Unfortunately, the sampling operation~$z {\sim} p_\P(z)$ is not differentiable.
Inspired by variational autoencoders~\cite{kingma2013auto}, we overcome this challenge via a reparameterization trick.
Accordingly, we set the opacity $\density$ of a Gaussian to be equal to the probability of the bucket that it was sampled from.
Consider that we sampled~$z {\sim} p_\P(z)$.
We then set $\density = \P_z$, i.e., we set $\density$ to the $z$-th entry of the vector of probabilities $\P$.
This means that in each backward pass, we assign the gradients of the loss $\mathcal{L}$ with respect to the opacities $\density$ to the gradients of the depth probability buckets $\depthprobs$, i.e., $\nabla_{\P} \loss = \nabla_{\density} \loss$.

To understand this approach at an intuitive level, consider the case where sampling produces a correct depth.
In this case, gradient descent \emph{increases} the opacity of the Gaussian, leading it to be sampled more often.
This eventually concentrates all probability mass in the correct bucket, creating a perfectly opaque surface.
Now, consider the case of an incorrectly sampled depth.
In this case, gradient descent \emph{decreases} the opacity of the Gaussian, lowering the probability of further incorrect depth predictions.

\paragraph{Predicting $\cov$ and $\sphhar$}
We predict a single covariance matrix and set of spherical harmonics coefficients per pixel by extending the neural network $f$ as
\begin{equation}
\P, \offset, \cov, \sphhar = f(\featmap[\pixcoord]).
\end{equation}

\paragraph{Summary}
Algorithm~\ref{alg} provides a summary of the procedure that predicts the parameters $(\mean, \cov, \density, \sphhar)$ of a pixel-aligned Gaussian primitive from the corresponding pixel's feature $\featmap[\pixcoord]$.

\begin{table*}[t]
\centering

\small

\resizebox{\textwidth}{!}{

\begin{tabular}{l|rrr|rrr|rr|rr}
\multicolumn{1}{l|}{}               &     \multicolumn{3}{c|}{ACID}                     &                   \multicolumn{3}{c|}{RealEstate10k}                 &      \multicolumn{2}{c|}{Inference Time (s)} & \multicolumn{2}{c}{Memory (GB)}   \\
                                    & PSNR $\uparrow$   & SSIM $\uparrow$   & LPIPS $\downarrow$ & PSNR $\uparrow$   & SSIM $\uparrow$   & LPIPS $\downarrow$  &     Encode $\downarrow$       & Render $\downarrow$ & Training $\downarrow$  & Inference $\downarrow$ \\                           
\midrule                                                                                                                                                                                                          
Ours                                &    \textbf{28.27} &    \textbf{0.843} &     \textbf{0.146} &    \textbf{26.09} &    \textbf{0.863} &     \textbf{0.136}  &                         0.102 &             \textbf{0.002} &              \textbf{14.4} &      \textbf{3.002} \\                                      
Du et al.~\cite{du2023cross}        & \underline{26.88} & \underline{0.799} &  \underline{0.218} & \underline{24.78} & \underline{0.820} &  \underline{0.213}  &                         0.016 &          \underline{1.309} &           \underline{314.3} &              19.604 \\                                      
GPNR~\cite{suhail2022generalizable} &             25.28 &             0.764 &              0.332 &             24.11 &             0.793 &              0.255  &                \textbf{N/A} &                     13.340 &                      3789.9 &              19.441 \\                                      
pixelNeRF~\cite{pixelnerf}          &             20.97 &             0.547 &              0.533 &             20.43 &             0.589 &              0.550  &             \underline{0.005} &                      5.294 &                       436.7 &   \underline{3.962} \\                                      
\end{tabular}

}

\caption{
\textbf{Quantitative comparisons.} We outperform all baseline methods in terms PSNR, LPIPS, and SSIM for novel view synthesis on the real-world RealEstate10k and ACID datasets. In addition, our method requires less memory during both inference and training and renders images about 650 times faster than the next-fastest baseline. In the memory column, we report memory usage for a single scene and $256 \times 256$ rays, extrapolating from the smaller number of rays per batch used to train the baselines where necessary. Note that we report GPNR's encoding time as N/A because it has no encoder. We \textbf{bold first-place} results and \underline{underline second-place} results in each column.}
\vspace{-5pt}
\label{tab:comparison_acid}
\end{table*}

\section{Experiments}
\label{sec:exp}

In this section, we describe our experimental setup, evaluate our method on wide-baseline novel view synthesis from image pairs, and perform ablations to validate our design.

\subsection{Experimental Setup}
\label{sec:experimental-setup}

We train and evaluate our method on RealEstate10k~\cite{realestate10k}, a dataset of home walkthrough videos downloaded from YouTube, as well as ACID~\cite{infinite_nature_2020}, a dataset of aerial landscape videos.
Both datasets include camera poses computed by SfM software, necessitating the scale-aware design discussed in~\Cref{sec:method-image-encoding}.
We use the provided training and testing splits. %
Because the prior state-of-the-art wide-baseline novel view synthesis model by Du et al.~\cite{du2023cross} only supports a resolution of $256 \times 256$, we train and evaluate our model at this resolution.
We evaluate our model on its ability to reconstruct video frames between two frames chosen as reference views.

\paragraph{Baselines} 
We compare our method against three novel-view-synthesis baselines.
pixelNeRF~\cite{pixelnerf} conditions neural radiance fields on 2D image features.
Generalizable Patch-based Neural Rendering (GPNR)~\cite{suhail2022generalizable} is an image-based light field rendering method that computes novel views by aggregating transformer tokens sampled along epipolar lines.
The unnamed method of Du et al.~\cite{du2023cross} also combines light field rendering with an epipolar transformer, but additionally uses a multi-view self-attention encoder and proposes a more efficient approach for sampling along epipolar lines.
\quad
To present a fair comparison, we retrained these baselines by combining their publicly available codebases with our datasets and our method's data loaders.
We train all methods, including ours, using the same training curriculum, where we gradually increase the inter-frame distance between reference views as training progresses.
For further training details, consult the supplementary material.

\paragraph{Evaluation Metrics} 
To evaluate visual fidelity, we compare each method's rendered images to the corresponding ground-truth frames by computing a peak signal-to-noise ratio (PSNR),  structural similarity index (SSIM)~\cite{Wang2004Image}, and  perceptual distance (LPIPS)~\cite{Zhang2018Unreasonable}.
We further evaluate each method's resource demands.
In this comparison, we distinguish between the encoding time, which is incurred once per scene and amortized over rendered views, and decoding time, which is incurred once per frame.

\paragraph{Implementation details}
Each reference image is passed through a ResNet-50~\cite{he2016deep} and a ViT-B/8 vision transformer~\cite{dosovitskiy2020image} that have both been pre-trained using a DINO objective~\cite{caron2021emerging}; we sum their pixel-wise outputs.
We train our model to minimize a combination of MSE and LPIPS losses using the Adam optimizer~\cite{DBLP:journals/corr/KingmaB14}. For the ``Plus Depth Regularization'' ablation, we regularize depth maps by fine-tuning with 50,000 steps of edge-aware total variation regularization.
Our encoder performs two rounds of epipolar cross-attention.

\subsection{Results}
\label{sec:results}

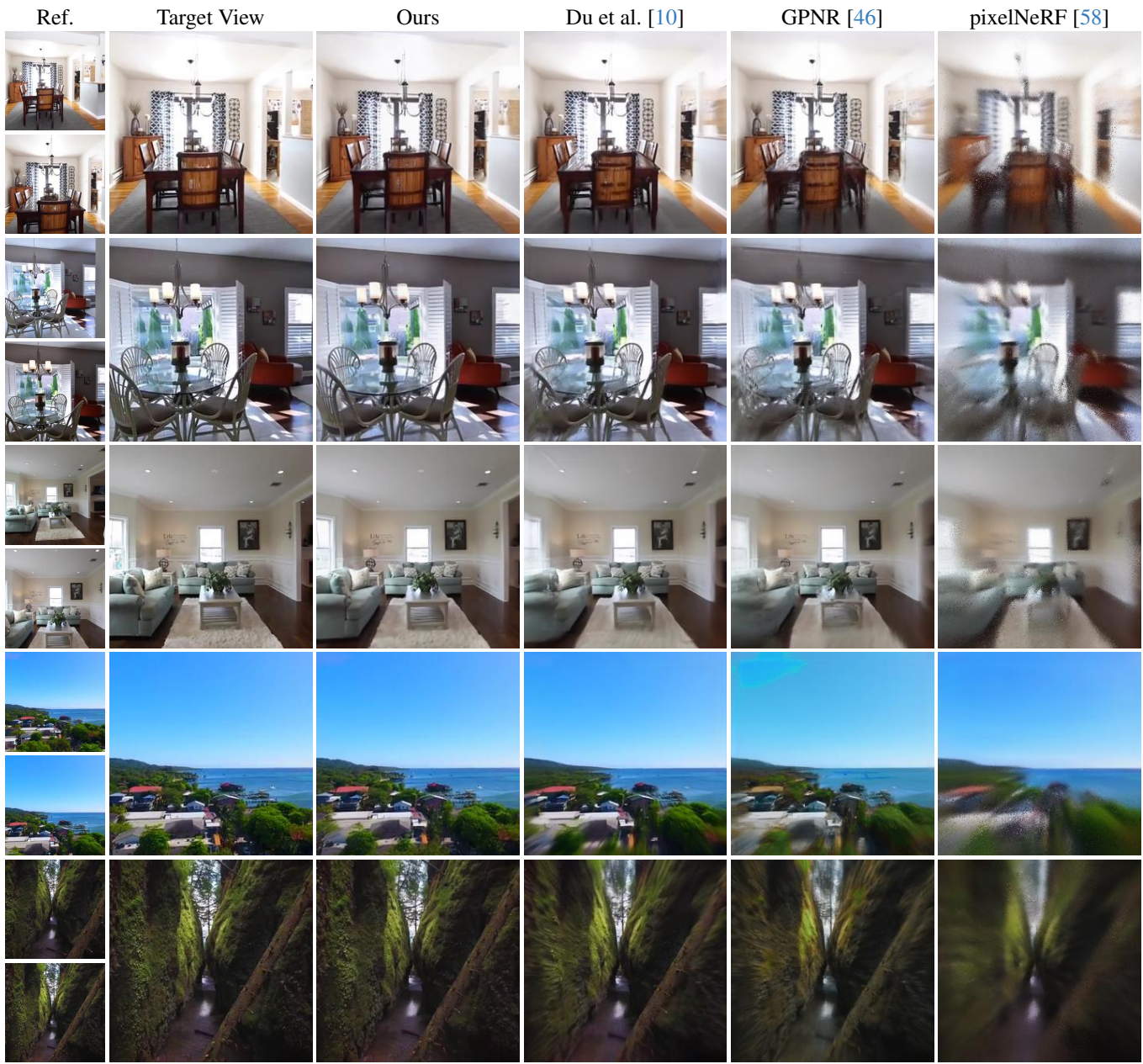
\begin{figure*}[t]
    \centering
    \def\svgwidth{1.02961\textwidth}
    \begingroup%
  \makeatletter%
  \providecommand\color[2][]{%
    \errmessage{(Inkscape) Color is used for the text in Inkscape, but the package 'color.sty' is not loaded}%
    \renewcommand\color[2][]{}%
  }%
  \providecommand\transparent[1]{%
    \errmessage{(Inkscape) Transparency is used (non-zero) for the text in Inkscape, but the package 'transparent.sty' is not loaded}%
    \renewcommand\transparent[1]{}%
  }%
  \providecommand\rotatebox[2]{#2}%
  \newcommand*\fsize{\dimexpr\f@size pt\relax}%
  \newcommand*\lineheight[1]{\fontsize{\fsize}{#1\fsize}\selectfont}%
  \ifx\svgwidth\undefined%
    \setlength{\unitlength}{375bp}%
    \ifx\svgscale\undefined%
      \relax%
    \else%
      \setlength{\unitlength}{\unitlength * \real{\svgscale}}%
    \fi%
  \else%
    \setlength{\unitlength}{\svgwidth}%
  \fi%
  \global\let\svgwidth\undefined%
  \global\let\svgscale\undefined%
  \makeatother%
  \begin{picture}(1,0.92490912)%
    \lineheight{1}%
    \setlength\tabcolsep{0pt}%
    \put(0,0){\includegraphics[width=\unitlength,page=1]{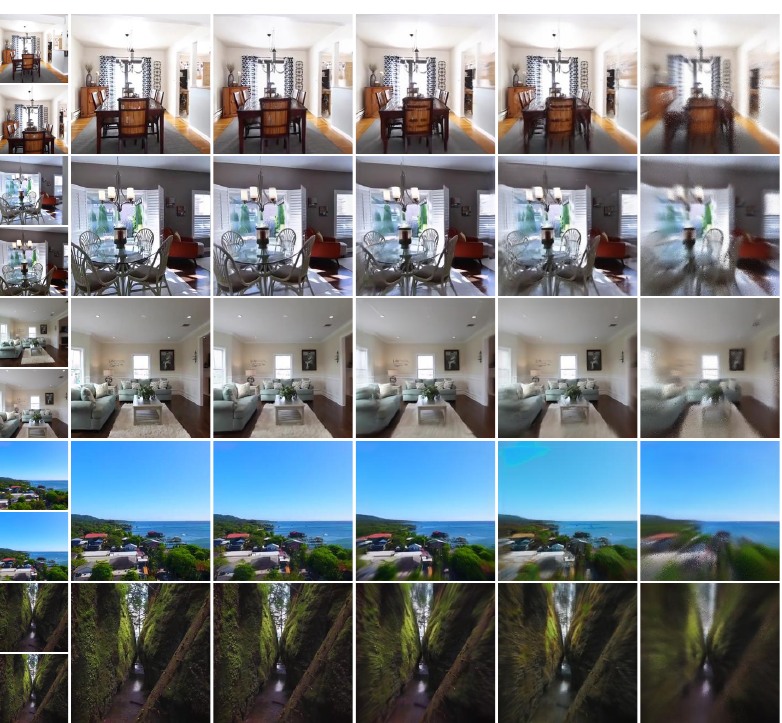}}%
    \put(0.04354546,0.91320912){\makebox(0,0)[t]{\lineheight{1.25}\smash{\begin{tabular}[t]{c}Ref.\end{tabular}}}}%
    \put(0.18018182,0.91320912){\makebox(0,0)[t]{\lineheight{1.25}\smash{\begin{tabular}[t]{c}Target View\end{tabular}}}}%
    \put(0.36236365,0.91320912){\makebox(0,0)[t]{\lineheight{1.25}\smash{\begin{tabular}[t]{c}Ours\end{tabular}}}}%
    \put(0.54454547,0.91320912){\makebox(0,0)[t]{\lineheight{1.25}\smash{\begin{tabular}[t]{c}Du et al.~\cite{du2023cross}\end{tabular}}}}%
    \put(0.72672729,0.91320912){\makebox(0,0)[t]{\lineheight{1.25}\smash{\begin{tabular}[t]{c}GPNR~\cite{suhail2022generalizable}\end{tabular}}}}%
    \put(0.90890912,0.91320912){\makebox(0,0)[t]{\lineheight{1.25}\smash{\begin{tabular}[t]{c}pixelNeRF~\cite{pixelnerf}\end{tabular}}}}%
  \end{picture}%
\endgroup%

    \vspace{-10pt}
    \caption{\textbf{Qualitative comparison of novel views on the RealEstate10k (top) and ACID (bottom) test sets.} Compared to the baselines, our approach not only produces more accurate and perceptually appealing images, but also generalizes better to out-of-distribution examples like the creek in the bottom row.}
    \label{fig:comparison_joint}
    \vspace{-1em}
\end{figure*}

We report quantitative results in~\Cref{tab:comparison_acid}.
Our method outperforms the baselines on all metrics, with especially significant improvements in perceptual distance (LPIPS).
We show qualitative results in Fig.~\ref{fig:comparison_joint}.
Compared to the baselines, our method is better at capturing fine details and correctly inferring 3D structure in portions of each scene that are only observed by one reference view.

\paragraph{Training and Inference cost}
As shown in \Cref{tab:comparison_acid}, our method is \textit{significantly} less resource-intensive than the baselines.
Compared to the next-fastest one, our method's cost to infer a single scene (encoding) and then render 100 images (decoding), the approximate number in a RealEstate10k or ACID sequence, is about 650 times less.
Our method also uses significantly less memory per ray at training time.

\paragraph{Point cloud rendering}
To qualitatively evaluate our method's ability to infer a structured 3D representation, we visualize the Gaussians it produces from views that are far outside the training distribution in~\Cref{fig:point_clouds}.
We visualize point clouds using the version of our model that has been fine-tuned with a depth regularizer.
Note that while the resulting Gaussians facilitate high-fidelity novel-view synthesis for in-distribution camera poses, they suffer from the same failure modes as 3D Gaussians optimized using the original 3D Gaussian splatting method~\cite{kerbl20233d}.
Specifically, reflective surfaces are often transparent, and Gaussians appear billboard-like when viewed from out-of-distribution views.

\subsection{Ablations and Analysis}
\label{sect:analysis}

\begin{figure}[!t]
    \centering
    \vspace{-15pt}
    \def\svgwidth{\linewidth}
    \begingroup%
  \makeatletter%
  \providecommand\color[2][]{%
    \errmessage{(Inkscape) Color is used for the text in Inkscape, but the package 'color.sty' is not loaded}%
    \renewcommand\color[2][]{}%
  }%
  \providecommand\transparent[1]{%
    \errmessage{(Inkscape) Transparency is used (non-zero) for the text in Inkscape, but the package 'transparent.sty' is not loaded}%
    \renewcommand\transparent[1]{}%
  }%
  \providecommand\rotatebox[2]{#2}%
  \newcommand*\fsize{\dimexpr\f@size pt\relax}%
  \newcommand*\lineheight[1]{\fontsize{\fsize}{#1\fsize}\selectfont}%
  \ifx\svgwidth\undefined%
    \setlength{\unitlength}{180bp}%
    \ifx\svgscale\undefined%
      \relax%
    \else%
      \setlength{\unitlength}{\unitlength * \real{\svgscale}}%
    \fi%
  \else%
    \setlength{\unitlength}{\svgwidth}%
  \fi%
  \global\let\svgwidth\undefined%
  \global\let\svgscale\undefined%
  \makeatother%
  \begin{picture}(1,0.78958333)%
    \lineheight{1}%
    \setlength\tabcolsep{0pt}%
    \put(0,0){\includegraphics[width=\unitlength,page=1]{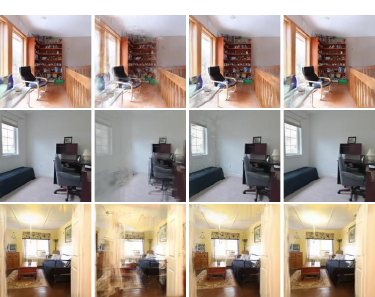}}%
    \put(0.121875,0.7625){\makebox(0,0)[t]{\lineheight{1.25}\smash{\begin{tabular}[t]{c}Ours\end{tabular}}}}%
    \put(0.37395833,0.7625){\makebox(0,0)[t]{\lineheight{1.25}\smash{\begin{tabular}[t]{c}No Epipolar\end{tabular}}}}%
    \put(0.62604167,0.7625){\makebox(0,0)[t]{\lineheight{1.25}\smash{\begin{tabular}[t]{c}No Sampling\end{tabular}}}}%
    \put(0.878125,0.7625){\makebox(0,0)[t]{\lineheight{1.25}\smash{\begin{tabular}[t]{c}Depth Reg.\end{tabular}}}}%
  \end{picture}%
\endgroup%

    \caption{\textbf{Ablations.} Without the epipolar transformer, our model is unable to resolve scale ambiguity, leading to ghosting artifacts. Without our sampling approach, our model falls into local minima that manifest themselves as speckling artifacts. Regularizing our model's predicted depths minimally affects rendering quality.}
    \label{fig:ablation}
\end{figure}
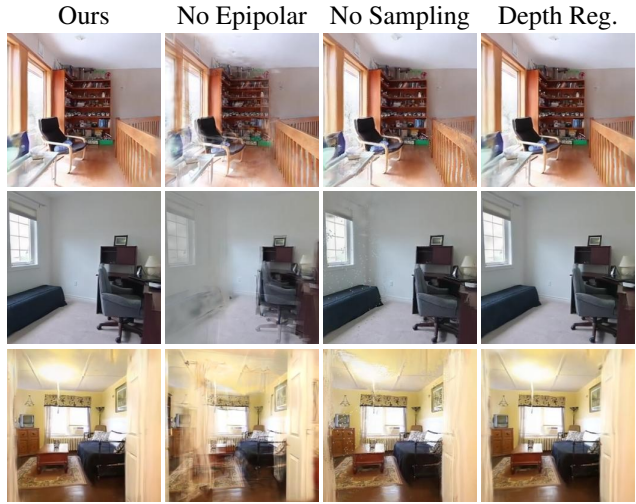

We perform ablations on RealEstate10k to answer the following questions:
\setdefaultleftmargin{.5em}{0em}{}{}{}{}
\begin{compactitem}
    \item \underline{Question 1a:} Is our epipolar encoder responsible for our model's ability to handle scale ambiguity?
    \item \underline{Question 1b:} If so, by what mechanism does our model handle scale ambiguity?
    \item \underline{Question 2:} Does our probabilistic primitive prediction alleviate local minima in primitive regression?
\end{compactitem}
See quantitative results in Tab.~\ref{tab:ablations} and qualitative ones in Fig.~\ref{fig:ablation}.

\begin{figure}[!t]
\centering
\includegraphics[width=\linewidth]{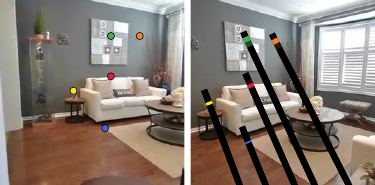}
\caption{\textbf{Attention visualization.} We visualize the epipolar cross-attention weights between the rays on the left and the corresponding epipolar lines on the right to confirm that our model learns to find the correct correspondence along each ray.}
\label{fig:attention}
\vspace{-15pt}
\end{figure}

\paragraph{Ablating epipolar encoding (Q1a)} 
To measure our epipolar encoding scheme's importance, we compare pixelSplat to a variant (No Epipolar Encoder) that eschews epipolar encoding.
Here, features are generated by encoding each reference view independently.
Qualitatively, this produces ghosting and motion blur artifacts that are evidence of incorrect depth predictions; quantitatively, performance drops significantly.
In~\Cref{fig:attention}, we visualize epipolar attention scores, demonstrating that our epipolar transformer successfully discovers cross-view correspondences.

\paragraph{Importance of depth for epipolar encoding (Q1b)} 
We investigate whether the frequency-based positional encoding of depth in Eq.~\ref{eq:pos_encoding} is necessary for our proposed epipolar layer.
We perform a study~(No Depth Encoding) where only image features $\featmap$ are fed to the epipolar attention layer. 
This leads to a performance drop of $\approx 1$dB PSNR.
This highlights that beyond simply detecting correspondence, our encoder uses the scene-scale encoded depths it triangulates to resolve scale ambiguity.

\paragraph{Importance of probabilistic prediction of Gaussian depths (Q2)}
To investigate whether predicting the depth of a Gaussian probabilistically is necessary, we perform an ablation (No Probabilistic Prediction) which directly regresses the depth, and thus the mean $\mathbf{\mu}$, of a pixel-aligned Gaussian with a neural network. 
We observe a performance drop of $\approx 1.5$dB in PSNR.

\begin{table}[t]
\centering

\small

\begin{tabular}{lrrr}
\toprule
Method        & PSNR $\uparrow$   & SSIM $\uparrow$   & LPIPS $\downarrow$ \\
\midrule
Ours          &    \textbf{26.09} &    \textbf{0.863} &     \textbf{0.136} \\
No Epipolar Encoder   &             19.89 &             0.639 &              0.295 \\
No Depth Encoding &             24.97 &             0.830 &              0.158 \\
No Probabilistic Sampling   &             24.62 &             0.828 &              0.171 \\
Plus Depth Regularization    & \underline{25.94} & \underline{0.862} &  \underline{0.137} \\
\bottomrule
\end{tabular}

\caption{ \textbf{Ablations.} Both our epipolar encoder and our probabilistic sampling scheme are essential for high-quality novel view synthesis. Depth regularization slightly impacts rendering quality.}
\vspace{-15pt}
\label{tab:ablations}
\end{table}

\section{Conclusion}
\label{sec:conclusion}
We have introduced pixelSplat, a method that reconstructs a primitive-based parameterization of the 3D radiance field of a scene from only two images.
At inference time, our method is significantly faster than prior work on generalizable novel view synthesis while producing an explicit 3D scene representation.
To solve the problem of local minima that arises in primitive-based function regression, we introduced a novel parameterization of primitive location via a dense probability distribution and introduced a novel reparameterization trick to backpropagate gradients into the parameters of this distribution. 
This framework is general, and we hope that our work inspires follow-up work on prior-based inference of primitive-based representations across applications.
An exciting avenue for future work is to leverage our model for generative modeling by combining it with diffusion models~\cite{tewari2023diffusion,szymanowicz2023viewset} or to remove the need for camera poses to enable large-scale training~\cite{smith2023flowcam}.

\paragraph{Acknowledgements} This work was supported by the National Science Foundation under Grant No. 2211259, by the Singapore DSTA under DST00OECI20300823 (New Representations for Vision), by the Intelligence Advanced Research Projects Activity (IARPA) via Department of Interior/ Interior Business Center (DOI/IBC) under 140D0423C0075, by the Amazon Science Hub, and by IBM. The Toyota Research Institute also partially supported this work. The views and conclusions contained herein reflect the opinions and conclusions of its authors and no other entity.

{
    \small
    \bibliographystyle{ieeenat_fullname}
    \bibliography{11_references}
}

\ifarxiv \clearpage \appendix \twocolumn[
    \centering
    \Large
    \textbf{\thetitle}\\
    \vspace{0.5em}Supplementary Material \\
    \vspace{1.0em}
] %

\section{Training Details}

We train all methods using our data loaders, which gradually increase the distance between reference views during training.
Specifically, over the first 150,000 training steps, we linearly increase the distance between reference views from 25 to 45.

\subsection{Our Method}

We train our method for 300,000 steps using a batch size of 7, which requires about 80 GB of VRAM on a single GPU. We use an MSE loss for the first 150,000 iterations and supplement it with an LPIPS loss with weight $0.05$ starting at 150,000 steps. For each batch element (scene), we render 4 target views. To allow our method to produce gradients for multiple estimated depths during each forward pass, we place 3 Gaussians along each ray and determine their positions by independently sampling from the ray's probability distribution 3 times. We then divide each Gaussian's alpha value by 3 such that $\alpha = \frac{\boldsymbol{\phi}_z}{3}$, which ensures that the Gaussians placed along any particular ray have a total opacity of roughly 1.
For an overview of pixelSplat's architecture, see Figure~\ref{fig:architecture}.

\begin{figure*}[!h]
\centering
\includegraphics[width=\textwidth]{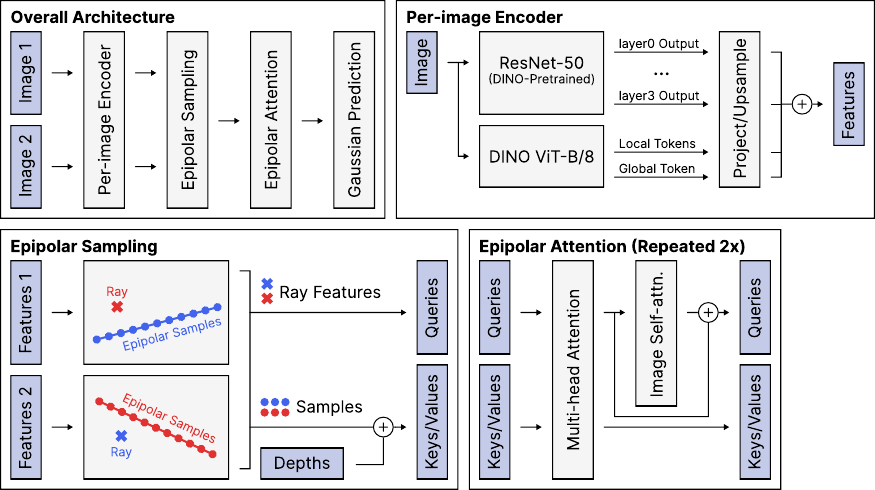}
\caption{\textbf{Architecture diagram.}}
\label{fig:architecture}
\end{figure*}

\paragraph{Depth regularization} To generate the point clouds shown in the main paper, we fine-tune our model for 50,000 steps with a depth regularization loss $\mathcal{L}_\text{reg. depth}$. To compute this loss, for each rendered view, we generate a corresponding depth map $D$. We use $D$ to compute the loss as follows, where $D[\pixcoord_x, \pixcoord_y]$ represents indexing:

{\footnotesize{
\begin{equation}
\begin{aligned}
    D^\Delta_x[\pixcoord] &= D[\pixcoord_{x - 1}, \pixcoord_y] - 2 D[\pixcoord_x, \pixcoord_y] + D[\pixcoord_{x + 1}, \pixcoord_y] \\
    D^\Delta_y[\pixcoord] &= D[\pixcoord_x, \pixcoord_{y - 1}] - 2 D[\pixcoord_x, \pixcoord_y] + D[\pixcoord_x, \pixcoord_{y + 1}] \\
    \img^\Delta_x &= \max(\left|\img[\pixcoord_{x - 1}, \pixcoord_y] - \img[\pixcoord_x, \pixcoord_y]\right|, \left|\img[\pixcoord_{x + 1}, \pixcoord_y] - \img[\pixcoord_x, \pixcoord_y]\right|) \\
    \img^\Delta_y &= \max(\left|\img[\pixcoord_x, \pixcoord_{y - 1}] - \img[\pixcoord_x, \pixcoord_y]\right|, \left|\img[\pixcoord_x, \pixcoord_{y + 1}] - \img[\pixcoord_x, \pixcoord_y]\right|) \\
    \mathcal{L}_\text{reg. depth}[\pixcoord] &= D^\Delta_x[\pixcoord] \exp\left(8 * \img^\Delta_x[\pixcoord] \right) + D^\Delta_y[\pixcoord] \exp\left(8 * \img^\Delta_y[\pixcoord] \right)
\end{aligned}
\end{equation}
}}

\paragraph{Transformation of Gaussians into world space}
Because our model predicts Gaussian parameters in camera space, these parameters must be transformed into world space before rendering.
Given a $4 \times 4$ camera-to-world extrinsics matrix $\mathbf{T}$ containing a $3 \times 3$ rotation $\mathbf{R}$ and a $3 \times 1$ translation $\mathbf{t}$, we transform them as follows:

\begin{equation}
\begin{aligned}
    \boldsymbol{\mu}_\text{world} &= \mathbf{T} \boldsymbol{\mu}_\text{cam.} \\
    \boldsymbol{\Sigma}_\text{world} &= \mathbf{R} \mathbf{\Sigma}_\text{cam.} \mathbf{R}^T \\
    \alpha_\text{world} &= \alpha_{cam.} \\
    \mathbf{s}_\text{world} &= \mathbf{D}_\mathbf{R} \mathbf{s}_\text{cam.}
\end{aligned}
\end{equation}

Here, $\mathbf{D}_\mathbf{R}$ is a block-diagonal matrix consisting of Wigner D matrices, which rotates the spherical harmonics coefficients $\mathbf{s}_\text{cam.}$.
In practice, we use the e3nn library to compute these matrices and recompile the original 3D Gaussian splatting code base to follow e3nn's conventions.

\subsection{Method of Du et al.}

We train the method of Du et al.~\cite{du2023cross} for 300,000 iterations using a total batch size of 32 spread across 4 GPUs, which requires around 44 GB of VRAM per GPU.
We train using the authors' default hyperparameters and enable the LPIPS loss after 150,000 iterations.

\subsection{pixelNeRF}

We train pixelNeRF~\cite{pixelnerf} for 500,000 iterations using a batch size of 12, which requires about 20 GB of VRAM on a single GPU.
We use the authors' default hyperparameters for the NMR dataset, in which the first pooling layer of pixelNeRF's ResNet is disabled to increase feature resolution.
Following Du et al.~\cite{du2023cross}, we set the near and far planes to be 0.1 and 10.0 respectively.

\subsection{GPNR}

We train GPNR~\cite{suhail2022generalizable} for 250,000 iterations using a batch size of 4098 spread across 6 GPUs, which requires about 67 GB of VRAM per GPU.
We use the authors' default hyperparameters but reduce the learning rate to $1 * 10^{-4}$, since several attempts at using the default learning rate of $3 * 10^{-4}$ yielded sudden training collapses on our dataset.

\section{Using More Reference Views}

While our epipolar encoder is primarily designed for novel view synthesis from pairs of images, it can be extended to an arbitrary number of views.
Specifically, for a given pixel coordinate, epipolar samples can be taken from any number of images.
The union of these samples can subsequently be used in place of a single epipolar line's samples.
To allow the epipolar transformer to distinguish between samples taken from different views, we add a learnable per-image embedding to each sampled feature.
Figure~\ref{fig:comparison_3_view} and Table~\ref{tab:three_view} show 3-view results.

\begin{figure*}[t]
    \centering
    \def\svgwidth{1.085\textwidth}
    \begingroup%
  \makeatletter%
  \providecommand\color[2][]{%
    \errmessage{(Inkscape) Color is used for the text in Inkscape, but the package 'color.sty' is not loaded}%
    \renewcommand\color[2][]{}%
  }%
  \providecommand\transparent[1]{%
    \errmessage{(Inkscape) Transparency is used (non-zero) for the text in Inkscape, but the package 'transparent.sty' is not loaded}%
    \renewcommand\transparent[1]{}%
  }%
  \providecommand\rotatebox[2]{#2}%
  \newcommand*\fsize{\dimexpr\f@size pt\relax}%
  \newcommand*\lineheight[1]{\fontsize{\fsize}{#1\fsize}\selectfont}%
  \ifx\svgwidth\undefined%
    \setlength{\unitlength}{375bp}%
    \ifx\svgscale\undefined%
      \relax%
    \else%
      \setlength{\unitlength}{\unitlength * \real{\svgscale}}%
    \fi%
  \else%
    \setlength{\unitlength}{\svgwidth}%
  \fi%
  \global\let\svgwidth\undefined%
  \global\let\svgscale\undefined%
  \makeatother%
  \begin{picture}(1,0.87285712)%
    \lineheight{1}%
    \setlength\tabcolsep{0pt}%
    \put(0,0){\includegraphics[width=\unitlength,page=1]{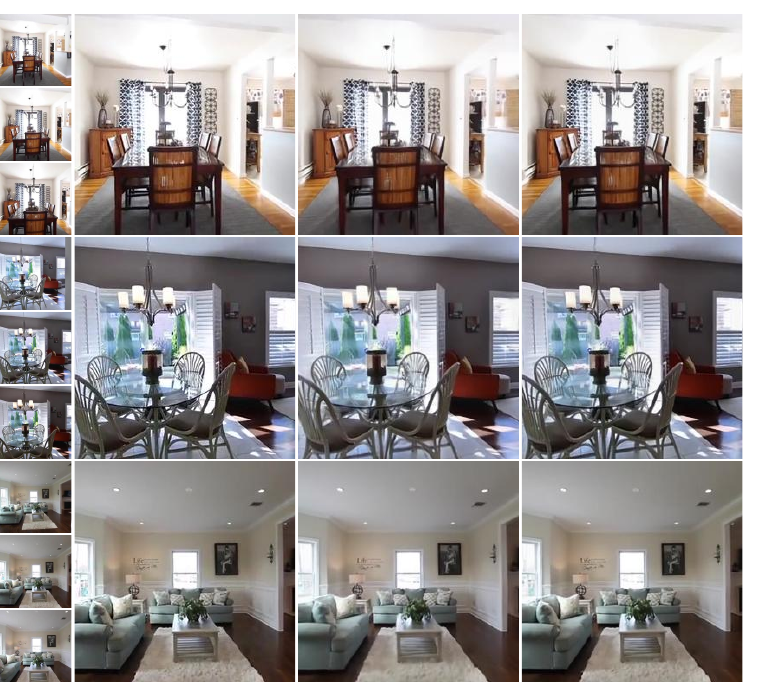}}%
    \put(0.04571429,0.86115712){\makebox(0,0)[t]{\lineheight{1.25}\smash{\begin{tabular}[t]{c}Ref.\end{tabular}}}}%
    \put(0.23657143,0.86115712){\makebox(0,0)[t]{\lineheight{1.25}\smash{\begin{tabular}[t]{c}Target View\end{tabular}}}}%
    \put(0.52285712,0.86115712){\makebox(0,0)[t]{\lineheight{1.25}\smash{\begin{tabular}[t]{c}Ours\end{tabular}}}}%
    \put(0.80914288,0.86115712){\makebox(0,0)[t]{\lineheight{1.25}\smash{\begin{tabular}[t]{c}Ours (3 Views)\end{tabular}}}}%
  \end{picture}%
\endgroup%

    \vspace{-10pt}
    \caption{\textbf{Qualitative comparison of novel views given 2 and 3 reference views.} Our method can be extended to use an arbitrary, fixed number of reference views as input. Here, we compare the results from a 3-view model with those from the original 2-view model. The 2-view model uses the top and bottom reference views as input, while the 3-view model uses all three.}
    \label{fig:comparison_3_view}
    \vspace{-1em}
\end{figure*}

\begin{table}
\centering

\footnotesize
\setlength{\tabcolsep}{2pt}

\begin{tabular}{lrrr}
\toprule
Method         & PSNR $\uparrow$   & SSIM $\uparrow$   & LPIPS $\downarrow$ \\
\midrule
Ours           & 26.09 & 0.863 &  0.136 \\
Ours (3 Views) &    28.31 &    0.908 &     0.100 \\
\bottomrule
\end{tabular}

\caption{
\textbf{Quantitative comparison of 2 and 3 view Real Estate 10k results.} Given a third reference view located halfway between the two existing reference views, a 3-view variant of pixelSplat produces slightly better results.}
\vspace{-1em}
\label{tab:three_view}
\end{table}

\section{Limitations}

Our model has several limitations.
First, rather than fusing or de-duplicating Gaussians observed from both reference views, it simply outputs the union of the Gaussians predicted from each view.
Second, it does not address generative modeling of unseen parts of the scene.
Finally, when extended to many reference views, our epipolar attention mechanism becomes prohibitively expensive in terms of memory.
Addressing these issues would be an exciting topic for future work.

\section{Additional Results}

We present additional results on the following pages.

\begin{figure*}[t]
    \centering
    \def\svgwidth{1.02961\textwidth}
    \vspace{-22pt}
    \begingroup%
  \makeatletter%
  \providecommand\color[2][]{%
    \errmessage{(Inkscape) Color is used for the text in Inkscape, but the package 'color.sty' is not loaded}%
    \renewcommand\color[2][]{}%
  }%
  \providecommand\transparent[1]{%
    \errmessage{(Inkscape) Transparency is used (non-zero) for the text in Inkscape, but the package 'transparent.sty' is not loaded}%
    \renewcommand\transparent[1]{}%
  }%
  \providecommand\rotatebox[2]{#2}%
  \newcommand*\fsize{\dimexpr\f@size pt\relax}%
  \newcommand*\lineheight[1]{\fontsize{\fsize}{#1\fsize}\selectfont}%
  \ifx\svgwidth\undefined%
    \setlength{\unitlength}{375bp}%
    \ifx\svgscale\undefined%
      \relax%
    \else%
      \setlength{\unitlength}{\unitlength * \real{\svgscale}}%
    \fi%
  \else%
    \setlength{\unitlength}{\svgwidth}%
  \fi%
  \global\let\svgwidth\undefined%
  \global\let\svgscale\undefined%
  \makeatother%
  \begin{picture}(1,1.28927271)%
    \lineheight{1}%
    \setlength\tabcolsep{0pt}%
    \put(0,0){\includegraphics[width=\unitlength,page=1]{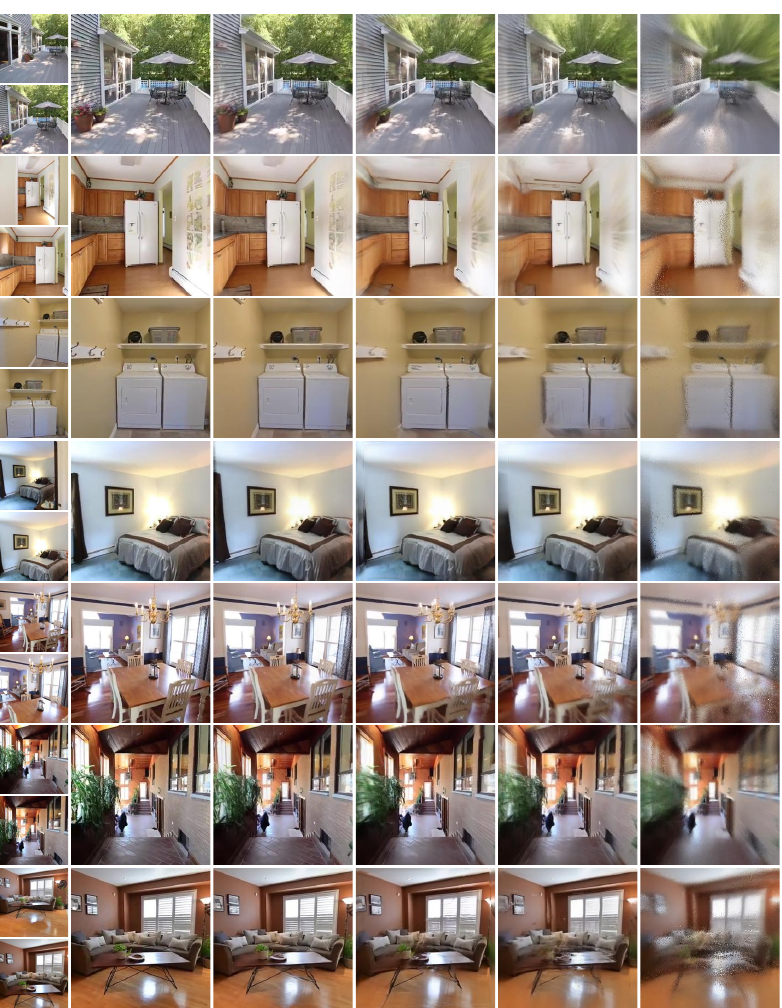}}%
    \put(0.04354546,1.27757271){\makebox(0,0)[t]{\lineheight{1.25}\smash{\begin{tabular}[t]{c}Ref.\end{tabular}}}}%
    \put(0.18018182,1.27757271){\makebox(0,0)[t]{\lineheight{1.25}\smash{\begin{tabular}[t]{c}Target View\end{tabular}}}}%
    \put(0.36236365,1.27757271){\makebox(0,0)[t]{\lineheight{1.25}\smash{\begin{tabular}[t]{c}Ours\end{tabular}}}}%
    \put(0.54454547,1.27757271){\makebox(0,0)[t]{\lineheight{1.25}\smash{\begin{tabular}[t]{c}Du et al.~\cite{du2023cross}\end{tabular}}}}%
    \put(0.72672729,1.27757271){\makebox(0,0)[t]{\lineheight{1.25}\smash{\begin{tabular}[t]{c}GPNR~\cite{suhail2022generalizable}\end{tabular}}}}%
    \put(0.90890912,1.27757271){\makebox(0,0)[t]{\lineheight{1.25}\smash{\begin{tabular}[t]{c}pixelNeRF~\cite{pixelnerf}\end{tabular}}}}%
  \end{picture}%
\endgroup%

    \vspace{-15pt}
    \caption{More results on the Real Estate 10k dataset.}
    \label{fig:more_comparison_re10k}
    \vspace{-1em}
\end{figure*}

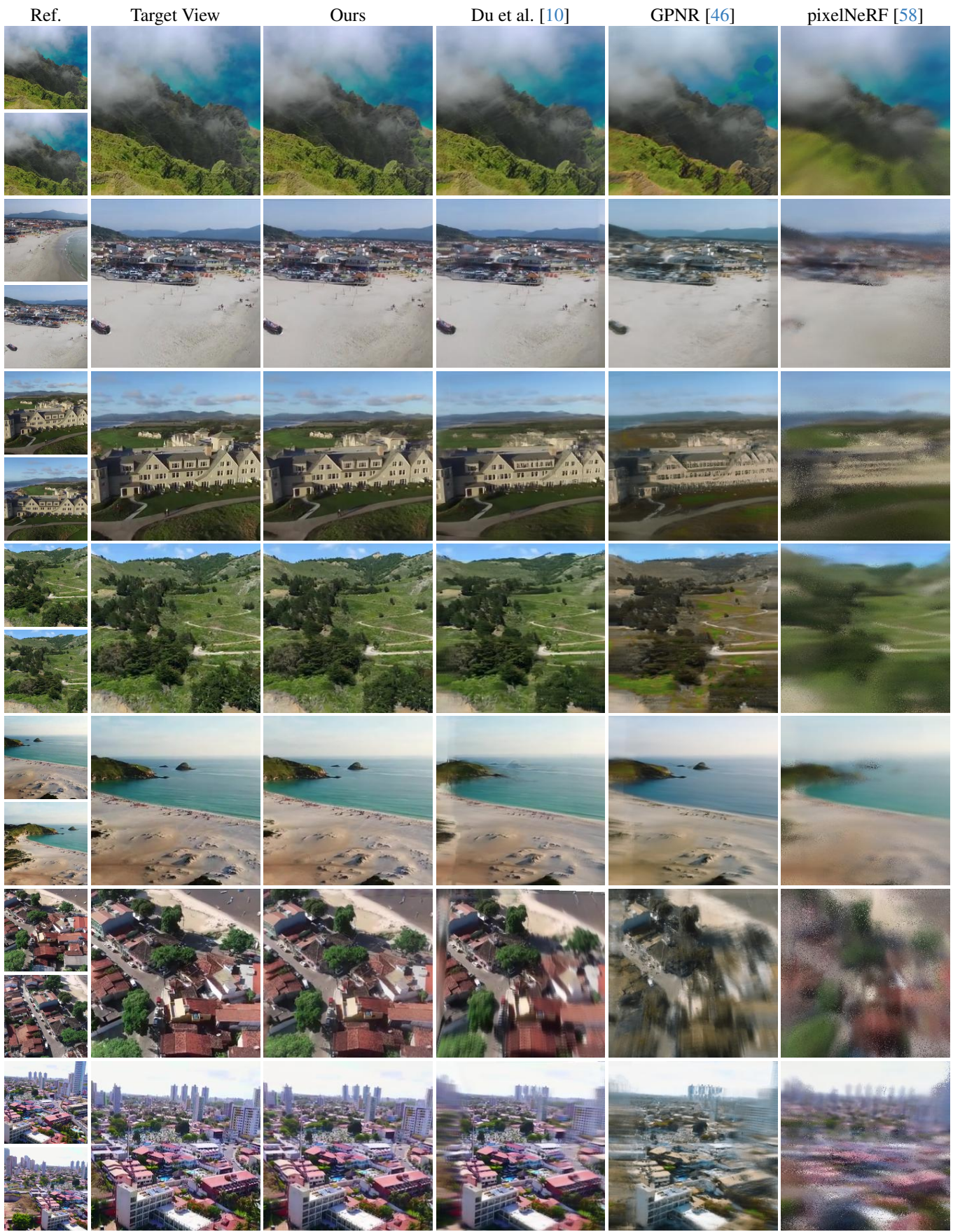
\begin{figure*}[t]
    \centering
    \def\svgwidth{1.02961\textwidth}
    \vspace{-22pt}
    \begingroup%
  \makeatletter%
  \providecommand\color[2][]{%
    \errmessage{(Inkscape) Color is used for the text in Inkscape, but the package 'color.sty' is not loaded}%
    \renewcommand\color[2][]{}%
  }%
  \providecommand\transparent[1]{%
    \errmessage{(Inkscape) Transparency is used (non-zero) for the text in Inkscape, but the package 'transparent.sty' is not loaded}%
    \renewcommand\transparent[1]{}%
  }%
  \providecommand\rotatebox[2]{#2}%
  \newcommand*\fsize{\dimexpr\f@size pt\relax}%
  \newcommand*\lineheight[1]{\fontsize{\fsize}{#1\fsize}\selectfont}%
  \ifx\svgwidth\undefined%
    \setlength{\unitlength}{375bp}%
    \ifx\svgscale\undefined%
      \relax%
    \else%
      \setlength{\unitlength}{\unitlength * \real{\svgscale}}%
    \fi%
  \else%
    \setlength{\unitlength}{\svgwidth}%
  \fi%
  \global\let\svgwidth\undefined%
  \global\let\svgscale\undefined%
  \makeatother%
  \begin{picture}(1,1.28927271)%
    \lineheight{1}%
    \setlength\tabcolsep{0pt}%
    \put(0,0){\includegraphics[width=\unitlength,page=1]{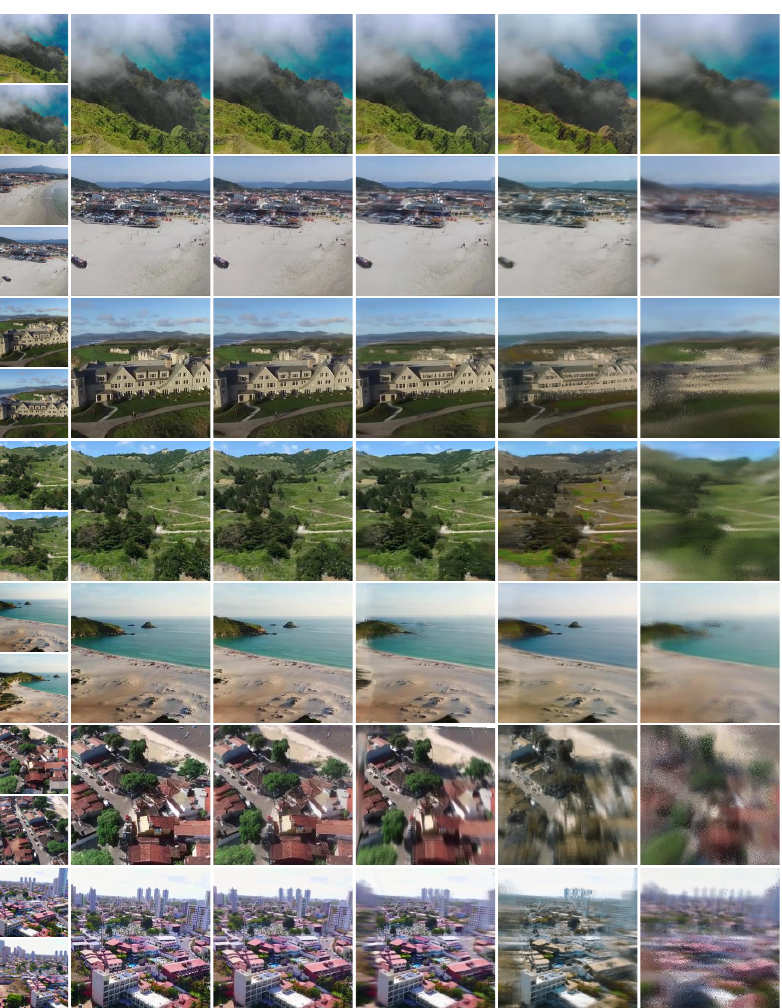}}%
    \put(0.04354546,1.27757271){\makebox(0,0)[t]{\lineheight{1.25}\smash{\begin{tabular}[t]{c}Ref.\end{tabular}}}}%
    \put(0.18018182,1.27757271){\makebox(0,0)[t]{\lineheight{1.25}\smash{\begin{tabular}[t]{c}Target View\end{tabular}}}}%
    \put(0.36236365,1.27757271){\makebox(0,0)[t]{\lineheight{1.25}\smash{\begin{tabular}[t]{c}Ours\end{tabular}}}}%
    \put(0.54454547,1.27757271){\makebox(0,0)[t]{\lineheight{1.25}\smash{\begin{tabular}[t]{c}Du et al.~\cite{du2023cross}\end{tabular}}}}%
    \put(0.72672729,1.27757271){\makebox(0,0)[t]{\lineheight{1.25}\smash{\begin{tabular}[t]{c}GPNR~\cite{suhail2022generalizable}\end{tabular}}}}%
    \put(0.90890912,1.27757271){\makebox(0,0)[t]{\lineheight{1.25}\smash{\begin{tabular}[t]{c}pixelNeRF~\cite{pixelnerf}\end{tabular}}}}%
  \end{picture}%
\endgroup%

    \vspace{-15pt}
    \caption{More results on the ACID dataset.}
    \label{fig:more_comparison_acid}
    \vspace{-1em}
\end{figure*}

 \fi

\end{document}